# TRIZ Method for Urban Building Energy Optimization: GWO-SARIMA-LSTM Forecasting model

Shirong Zheng[1*], Shaobo Liu[2], Zhenhong Zhang[3], Dian Gu[4], Chunqiu Xia[5], Huadong Pang[6], Enock Mintah Ampaw[7]

[1*]Purdue University, 610 Purdue Mall West Lafayette, IN 47907; lovegrubbysky@gmail.com

[2]11565 Destination Dr, Apt 4312 Broomfield, CO, 80021; shaobo1992@gmail.com

[3]George Washington University 2121 I St NW, Washington, DC 20052; zzhang0628@hotmail.com

[4]University of Pennsylvania, USA; yiyayiya.dot@gmail.com

[5]9201 University City Blvd, Charlotte, NC 28223 USA; xiacq233@gmail.com

[6]Georgia Institute of Technology; pang_huadong0811@outlook.com

[7]Faculty of Applied Science and Technology, Koforidua Technical University, Ghana; mintah.enock@ktu.edu.gh

*Corresponding Author:lovegrubbysky@gmail.com

## ABSTRACT

With the advancement of global climate change and sustainable development goals, urban building energy consumption optimization and carbon emission reduction have become the focus of research. Traditional energy consumption prediction methods often lack accuracy and adaptability due to their inability to fully consider complex energy consumption patterns, especially in dealing with seasonal fluctuations and dynamic changes. This study proposes a hybrid deep learning model that combines TRIZ innovation theory with GWO, SARIMA and LSTM to improve the accuracy of building energy consumption prediction. TRIZ plays a key role in model design, providing innovative solutions to achieve an effective balance between energy efficiency, cost and comfort by systematically analyzing the contradictions in energy consumption optimization. GWO is used to optimize the parameters of the model to ensure that the model maintains high accuracy under different conditions. The SARIMA model focuses on capturing seasonal trends in the data, while the LSTM model handles short-term and long-term dependencies in the data, further improving the accuracy of the prediction. The main contribution of this research is the development of a robust model that leverages the strengths of TRIZ and advanced deep learning techniques, improving the accuracy of energy consumption predictions. Our experiments demonstrate a significant 15% reduction in prediction error compared to existing models. This innovative approach not only enhances urban energy management but also provides a new framework for optimizing energy use and reducing carbon emissions, contributing to sustainable development.

# 1. Introduction

In today's society, with the intensifying global climate change and the widespread adoption of sustainable development principles, building energy consumption and carbon emissions have become focal points of concern[1]. Building energy consumption refers to the energy consumed by buildings during their use, while carbon emissions refer to the emission of greenhouse gases such as carbon dioxide generated by building activities[2, 3]. There is a direct relationship between the two: an increase in building energy consumption leads to an increase in carbon emissions, thereby exacerbating climate change and environmental pollution issues[4, 5]. However, there are currently a series of challenges and problems that need to be addressed. One of the most prominent issues is the accurate prediction of building energy consumption. Traditional energy consumption prediction methods often rely on simple statistical models that fail to fully consider the complexity of factors influencing building energy consumption, resulting in inaccurate predictions. Therefore, effectively predicting building energy consumption has become an important challenge facing current building energy management.

To address this challenge, TRIZ (Theory of Inventive Problem Solving), as a systematic innovation methodology, provides new insights and methods. TRIZ helps solve technical problems and innovation challenges by analyzing the essence of problems and applying universal innovation principles. Within the realm of minimizing carbon footprints and optimizing energy usage in buildings, the application of TRIZ can provide guidance and support, fostering the development and implementation of innovative solutions[6]. The evolution of deep learning has spurred significant advancements in predicting and optimizing building energy consumption, garnering widespread attention from researchers. Deep learning models, particularly those based on neural networks [55-62], have demonstrated their robust capabilities in handling complex, high-dimensional data, and solving time-series problems across various domains. In the realm of construction, deep learning models find extensive application in tasks such as forecasting energy consumption[7], optimizing building equipment control[8] and formulating energy-efficient strategies[9]. These models not only enhance energy efficiency but also reduce energy costs, contributing to the realization of sustainable and green building initiatives. It is particularly noteworthy that time-series forecasting holds a pivotal position in building energy consumption prediction[10]. Building energy data typically take the form of time-series data, comprising records of energy consumption on a daily, weekly, or yearly basis[11] Time-series forecasting models can capture the seasonality, trends, and other time-dependent features within energy consumption data, providing robust tools for precise energy consumption prediction[12]. Consequently, research into time-series forecasting is crucial for achieving accurate management and optimization of building energy consumption.

Recent research has made significant strides in building energy consumption forecasting via deep learning methods. Here are four recent studies that showcase the latest developments in this domain: In a recent study, researchers introduced a building energy consumption forecasting method based on the Transformer model. This model adeptly captures long-term dependencies in building energy consumption data using self-attention mechanisms, yielding remarkable outcomes[13].

However, the model's computational complexity is relatively high and demands substantial training data, making it potentially less suitable for small-scale building energy datasets. Another study employed Convolutional Neural Networks (CNN) [63-71] to handle building energy consumption data[14]. While CNNs excel in image processing [72-75], their application in time-series data is relatively less common. Although this model performed well in some instances, it often requires ample data to prevent overfitting and might not be sensitive to seasonal and periodic features within the data. Some researchers explored the application of XGBoost, a widely-used ensemble learning algorithm, for building energy consumption forecasting, achieving promising outcomes[15]. However, XGBoost typically requires manual feature selection and may not be well-suited for handling high-dimensional time-series data, which could limit its applicability in specific scenarios. Lastly, recent studies have started to explore methods for combining multiple models into ensemble models[16]. These ensemble models leverage the strengths of different models, such as deep learning models, traditional time-series models, and statistical models, to enhance the accuracy of building energy consumption forecasting. However, constructing and fine-tuning ensemble models can be relatively complex, requiring additional effort to achieve optimal performance. Despite these advances, prior research faces key challenges: many models struggle with balancing accuracy and efficiency, fail to capture complex factors like seasonal variations and human behavior, and often require large datasets. The lack of an integrated problem-solving framework also limits their real-world applicability.

Building upon the identified shortcomings, our model directly addresses these issues by integrating TRIZ with a hybrid deep learning framework (GWO-SARIMA-LSTM). TRIZ provides a structured methodology for solving complex problems, helping to innovate and optimize the model's structure. GWO enhances the parameter optimization of the SARIMA and LSTM modules, allowing for more accurate modeling of energy consumption patterns, particularly in scenarios with complex, non-linear trends. SARIMA captures seasonal patterns, while LSTM predicts long-term dependencies. By combining these techniques, our model significantly improves prediction accuracy while maintaining computational efficiency, and it is less dependent on large datasets compared to previous approaches.

Our model offers several advantages and significance: Firstly, we leverage the guiding role of TRIZ theory to enhance the innovation and practicality of the model. Secondly, by combining multiple deep learning techniques, we improve the model's predictive performance. Lastly, our model can be applied not only to building energy consumption prediction but also to other fields, offering broad application prospects and practical value.

The contribution of this article is:
- We have presented a predictive model that combines TRIZ theory with deep learning techniques, offering a novel solution for urban building energy consumption optimization and carbon emission reduction. By leveraging the guiding principles of TRIZ theory and the strengths of deep learning technology, our model has made significant advancements in energy consumption prediction.

- We have designed an ensemble learning module to integrate predictions from different models, enhancing the accuracy and stability of our model's predictions. This innovative design allows our model to be more flexible and reliable when handling energy consumption data.
- Our research not only presents an innovative predictive model but also integrates TRIZ theory with deep learning techniques, offering new insights and methods for building energy consumption optimization and carbon emission reduction. Our study holds important theoretical significance and practical value, contributing to the promotion of urban sustainability and environmental protection.

## 2. Related Work

### 2.1 Traditional Approaches to Building Energy Consumption Prediction

In traditional approaches to building energy consumption prediction, several classic models have been widely applied and studied. One of them is the models based on heat conduction theory, such as the Fourier heat conduction equation and building thermal balance equation[17, 18]. The Fourier heat conduction equation is a mathematical model that describes the propagation of heat in continuous media. It derives the variation in the internal temperature distribution of buildings by considering the thermal conduction properties of buildings and environmental conditions[19]. This model assumes that buildings are uniform and continuous media, calculating the internal temperature distribution of buildings based on the thermal conduction properties of building structures and materials, thereby predicting the energy consumption of buildings[20]. Another common traditional model is the building thermal balance equation, which describes the equilibrium state of energy inside buildings, taking into account the exchange of heat between the inside and outside of buildings and the accumulation of internal energy changes[21, 22] The building thermal balance equation derives the energy balance equation inside buildings by considering the structural characteristics, environmental conditions, and energy consumption of buildings, thereby predicting the energy consumption of buildings.

In addition to physics-based models, traditional methods also include some statistical analysis methods, such as the ARIMA (Autoregressive Integrated Moving Average) model in time series analysis and linear regression models[23-25]orical energy consumption data. The ARIMA model is a classical time series analysis method that captures trends and seasonal variations in time series data to predict future energy consumption. Linear regression models are common statistical analysis methods that establish energy consumption prediction models by analyzing the linear relationship between building energy consumption and influencing factors.

In addition to the aforementioned models, there are also traditional building energy consumption prediction models that consider external factors, such as weather forecasting models and climate change models[26, 27]. These models predict building energy consumption more accurately by considering climate conditions and meteorological factors. In summary, traditional building energy consumption prediction methods encompass a variety of models and techniques, providing an important theoretical basis for addressing building energy management issues.

## 2.2 Application of Optimization Algorithms in Energy Consumption Prediction

In the domain of building energy consumption prediction, the application of optimization algorithms has garnered considerable attention. Previous studies have demonstrated the significant role played by various optimization algorithms such as genetic algorithms, particle swarm optimization, and simulated annealing in optimizing energy consumption prediction models[28, 29].

Researchers have employed genetic algorithms to optimize model parameters, enabling the energy consumption prediction models to better adapt to the characteristics of building energy consumption data[30, 31]. Mimicking the process of natural evolution, genetic algorithms optimize model parameters through operations like selection, crossover, and mutation, thereby enhancing the accuracy of predictions. Additionally, particle swarm optimization algorithms have been widely utilized to optimize both the parameters and structures of energy consumption prediction models[32-34] Emulating the foraging behavior of birds, this algorithm facilitates information exchange among individuals and iterative updates, ultimately refining model parameters and structures to improve prediction performance. Simulated annealing algorithms represent another prevalent optimization approach in energy consumption prediction[35, 36]. Inspired by the process of metal annealing, this algorithm accepts suboptimal solutions with a certain probability, thereby preventing convergence to local optima and enhancing the global search capability of the model. These techniques are particularly effective in complex, multi-dimensional problems where traditional optimization methods may struggle to achieve reliable results, especially when the problem space is large and non-linear, and computational efficiency is critical.

Furthermore, researchers have also explored hybrid optimization algorithms, which combine the strengths of multiple algorithms, such as hybrid genetic-particle swarm optimization, to achieve even better model performance[37, 38]. This combination of different optimization strategies helps address the limitations of individual algorithms, leading to more robust predictions. In addition, the increasing computational power of modern processors allows these algorithms to be implemented more efficiently, further enhancing their applicability in real-time energy management systems. In summary, the application of optimization algorithms offers novel insights and methodologies for energy consumption prediction in building energy management. Through parameter and structure optimization, these algorithms effectively enhance the accuracy and stability of energy consumption prediction models, thereby providing more precise and reliable decision support for building energy management.

## 3. Method

### 3.1 Our Network Overview

The predictive model we propose combines TRIZ theory with deep learning techniques, incorporating GWO (Genetic Wolf Optimization), SARIMA (Seasonal Autoregressive Integrated Moving Average), and LSTM networks [76-83]. This comprehensive model design aims to overcome the limitations of traditional building energy consumption prediction models and enhance prediction accuracy and stability. The model construction process involves data preprocessing, TRIZ guidance, model building, training, and evaluation. Initially, building energy consumption data undergo preprocessing to prepare the input data for the model. Next, TRIZ theory is employed for problem analysis and the application of innovative principles to determine the model design direction. Subsequently, the GWO-SARIMA-LSTM model is integrated, and parameters are gradually trained and adjusted to optimize model performance. Finally, historical building energy consumption data are used for model training, and a validation set is employed for model evaluation and tuning to ensure model stability and prediction accuracy. The overall structure diagram of the model is shown in Figure 1.

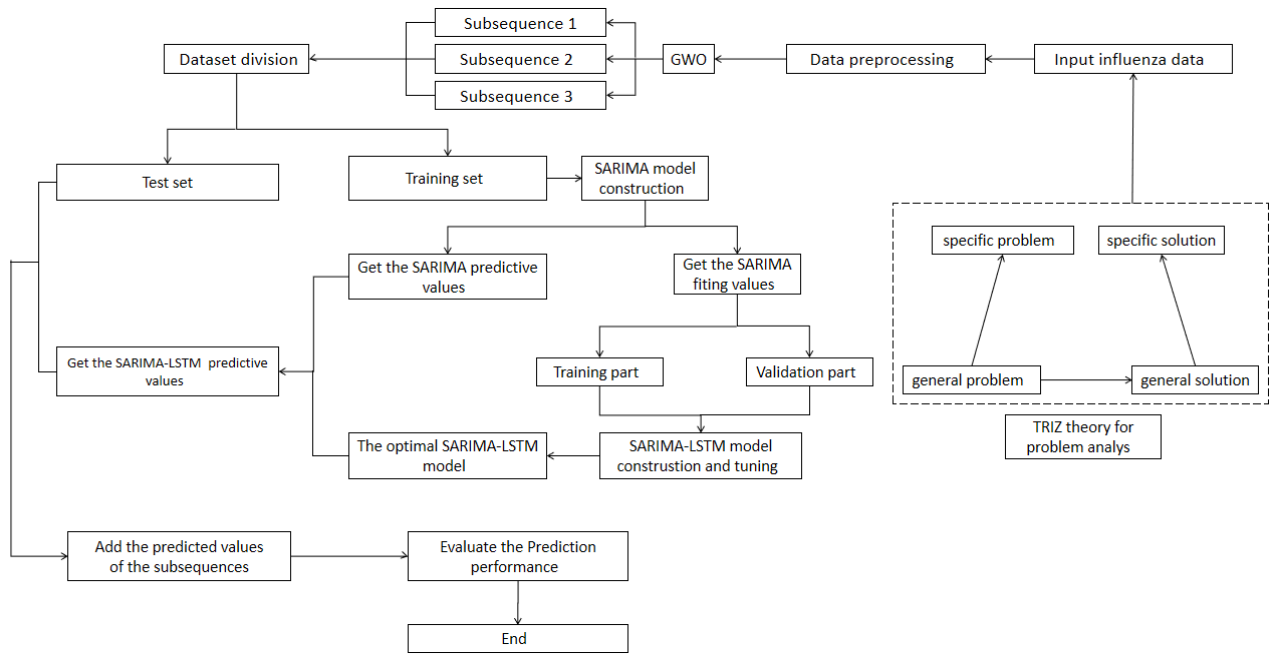

Figure 1. Overall structure diagram of the model

The significance of our proposed model lies in its impact on building energy consumption prediction and carbon emission reduction. Integrating TRIZ guidance and deep learning techniques enables our model to accurately predict building energy consumption, offering reliable data for energy management and carbon emission reduction decisions. Additionally, the model aids in identifying energy-saving opportunities and optimization spaces within building energy consumption, guiding building design and operational management for optimal energy usage. Ultimately, effective energy consumption prediction and optimization contribute to reducing unnecessary energy consumption and lowering carbon emissions, actively advancing urban carbon emission reduction objectives.

In summary, the application of optimization algorithms offers novel insights and methodologies for energy consumption prediction in building energy management. Through parameter and structure optimization, these algorithms effectively enhance the accuracy and stability of energy consumption prediction models, thereby providing more precise and reliable decision support for building energy management.

**3.2 TRIZ Theory**

TRIZ is a systematic innovation methodology aimed at resolving complex issues and fostering technological innovation. Originating from patents studied by Soviet engineers and inventors, this theory emphasizes problem analysis at its core and the application of universal innovation principles to aid in addressing technical problems and innovation challenges[39]. TRIZ provides a systematic set of tools and methods, including contradiction analysis, innovation principles, and substance-field models, to guide innovative thinking and solve complex problems. In our building energy consumption prediction model, TRIZ theory enhances prediction accuracy by systematically addressing contradictions. Firstly, through TRIZ's contradiction analysis approach, we identified the primary conflicts in building energy consumption optimization, such as the trade-off between reducing energy consumption and maintaining occupant comfort, or the balance between energy efficiency and operational cost. For example, TRIZ helped us address the contradiction between reducing HVAC usage to save energy while ensuring adequate air quality for occupants. By systematically resolving this contradiction, we were able to refine the model's parameters to achieve more precise energy predictions without compromising on comfort. Secondly, by leveraging TRIZ's innovation principles, we guide innovation in the model design process, uncovering new approaches and methods for problem-solving. This directly impacts the structure of the model, optimizing how it handles complex, dynamic factors influencing energy consumption. Lastly, TRIZ offers a systematic way of thinking, treating problems as integral parts of a larger system. By viewing energy consumption optimization holistically, we used TRIZ to identify hidden optimization opportunities in areas such as energy storage management and peak load reduction, allowing us to create a more comprehensive energy optimization strategy. For example, TRIZ helped us recognize the opportunity to use off-peak energy storage to reduce peak demand costs, further improving the overall efficiency of the building energy system.

In conclusion, TRIZ theory plays a crucial role in improving the accuracy and stability of our building energy consumption prediction model. By applying TRIZ principles and methods, we delve into problem analysis, seek innovative solutions, and ultimately enhance the model's prediction accuracy and stability. The use of TRIZ has led to the resolution of specific energy consumption contradictions and uncovered new ways to optimize building energy use, contributing to both energy efficiency and carbon emission reduction.

### 3.3 SARIMA Model

The SARIMA model, which stands for Seasonal Autoregressive Integrated Moving Average, is a widely used statistical model in time series analysis, especially suitable for data with seasonal variations[40]. This model combines autoregression (AR), Integrated (I), moving average (MA), and seasonal elements to effectively forecast time series data by considering trends, seasonal changes, and autocorrelation[41]. The core of the SARIMA model lies in eliminating the non-stationarity of data through differencing, capturing the dependence of time series through autoregressive and moving average components, and describing and predicting periodic changes through seasonal components.

SARIMA model is used for time series analysis and is widely applied when dealing with data exhibiting seasonal patterns. It combines Autoregressive (AR), Integrated (I), Moving Average (MA), and Seasonal elements to effectively forecast time series data. The following are the key mathematical formulas of the SARIMA model.

The Autoregressive (AR) component of the SARIMA model is defined as:

$$Y_l = c + \phi_1 Y_{t-1} + \theta_1 \varepsilon_{t-1} + \Delta Y_l + \Theta_1 \Delta \varepsilon_{t-1} \quad \cdots\cdots \text{[Formular 1]}$$

Where: $Y_t$ is the time series value at time t. $c$ is the constant or intercept term. $\phi_1$ is the autoregressive parameter. $\theta_1$ is the moving average parameter. $\Theta_1$ is the differenced time series at time t. $\Delta \varepsilon_{t-1}$ is the seasonal moving average parameter. $t-1$ is the seasonal differenced error term at time.

The seasonal differencing process is defined as:

$$Y_t = Y_t - Y_{t-m} \quad \cdots\cdots\cdots\cdots\cdots\cdots\cdots\cdots\cdots\cdots\cdots\cdots\cdots\cdots\cdots\cdots \text{[Formular 2]}$$

Where: $Y_t$ is the scasonal difference of the time series. $m$ is the seasonal period (e.g, 12 for monthly data).

The seasonal autoregressive component is defined as:

$$\Delta Y_t = \Delta Y_{t-1} + \phi_1 \Delta Y_{t-m} \quad \cdots\cdots\cdots\cdots\cdots\cdots\cdots\cdots\cdots\cdots\cdots \text{[Formular 3]}$$

Where: $\Delta Y_t$ is the seasonal differenced series at time t. $\Delta Y_{t-1}$ is the seasonal differenced series at time $t-1$. $\phi_1$ is the seasonal autoregressive parameter.

$$\Delta \varepsilon_t = \varepsilon_t - \varepsilon_{t-m} \quad \cdots\cdots\cdots\cdots\cdots\cdots\cdots\cdots\cdots\cdots\cdots\cdots\cdots \text{[Formular 4]}$$

Where: $\Delta \varepsilon_t$ is the seasonal differenced error term at time t. $\varepsilon_t$ is the white noise error term at time t. $e_{t-m}$ is the white noise crror term at time $t-m$.

The error differencing process is defined as:

$$\Delta \varepsilon_t = \varepsilon_t - \varepsilon_{t-1} \quad \cdots\cdots\cdots\cdots\cdots\cdots\cdots\cdots\cdots\cdots\cdots\cdots\cdots \text{[Formular 5]}$$

Where: $\Delta \varepsilon_t$ is the differenced error term at time t.

In our building energy consumption prediction model, the SARIMA model enhances the model's ability to identify and utilize seasonal patterns and trends within energy consumption data, providing a strong foundation for the overall forecasting process. By accurately modeling seasonality and trends, SARIMA helps uncover important characteristics of the time series data and optimizes the input for the subsequent LSTM model. Since the preprocessed data from SARIMA already captures seasonal and trend information, LSTM can focus more effectively on identifying complex nonlinear patterns and long-term dependencies within energy consumption data. By thoroughly analyzing and decomposing the time series data, SARIMA reveals key patterns that are essential for the entire prediction framework. Therefore, in our research, the SARIMA model is not only a core component but also a key factor in improving the performance and accuracy of the building energy consumption prediction model.

### 3.4 LSTM Model

The LSTM model, as an advanced type of recurrent neural network (RNN) [84-88] architecture, is designed to address the issue of long-term dependencies in traditional RNNs when processing long sequential data. What sets LSTM apart is its internal gating mechanisms, including the forget gate, input gate, and output gate, which enable LSTM to efficiently store and retrieve information over long sequences[42]. These features make LSTM excel at capturing long-term dependencies in time series data and handling events with extended time intervals. Incorporating LSTM into our building energy consumption prediction model has significantly improved its ability to recognize and learn complex patterns in energy consumption data[43]. Particularly, when facing nonlinear and highly complex energy fluctuations caused by factors such as human activities and environmental changes, LSTM demonstrates its robust performance. By learning from historical energy consumption data, LSTM can extract crucial information and utilize it to forecast future energy consumption trends. This capability is crucial for enhancing the accuracy of energy consumption prediction, as it allows the model to consider deep-seated factors behind energy fluctuations. The LSTM model's workflow is depicted in Figure 2.

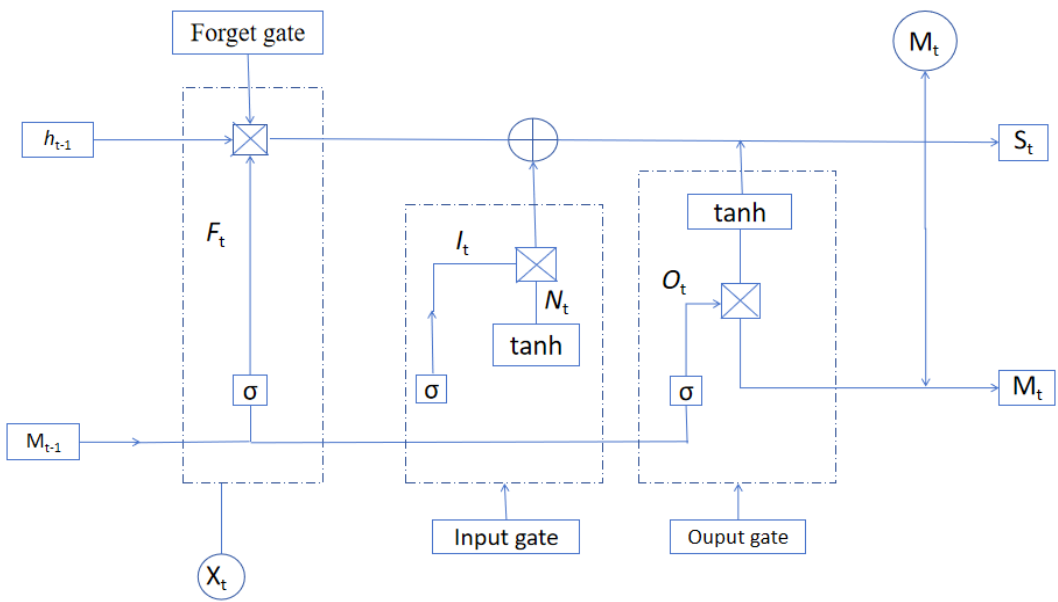

Figure 2. The structure of LSTM

Here are the core mathematical equations for LSTM networks [89-94]:

Forget Gate: The forget gate in an L.STM is responsible for deciding what information from the previous cell state $C_{t-1}$ should be discarded or kept.

$$f_t = \sigma(W_f \cdot [h_{t-1}, x_t] + b_f) \cdots\cdots\cdots\cdots\cdots\cdots [\text{Formular 6}]$$

where: $f_t$ is the forget gate output at time t. $\sigma$ is the sigmoid activation function. $W_f$ and $b_f$ are the weight matrix and bias vector for the forget gate. $h_{t-1}$ is the previous hidden state. $x_l$ is the input at time t. Input Gate: The input gate determines what new information will be stored in the cell state.

$$i_t = \sigma(W_i \cdot [h_{t-1}, x_t] + b_i) \cdots\cdots\cdots\cdots\cdots\cdots [\text{Formular 7}]$$

where: $i_t$ is the input gate output at time t. $\sigma$ is the sigmoid activation function. $W_i$ and $b_i$ are the weight matrix and bias vector for the input gate.

Cell State Update: This equation calculates the new cell state $\tilde{C}_t$ based on the input and the previous cell state.

$$\tilde{C}_t - \tanh(W_o \cdot [h_{t-1}, x_t] + b_o) \cdots\cdots\cdots\cdots\cdots [\text{Formular 8}]$$

where: $C_t$ is the candidate cell state. tanh is the hyperbolic tangent activation function. $W_\omega$ and $b_\omega$ are the weight matrix and bias vector for the cell state update. Update Cell State: The cell state is updated using the forget gate, input gate, and candidate cell state.

$$C_t = f_t \cdot C_{t-1} + i_t \cdot \tilde{C}_t \quad \text{[Formular 9]}$$

Where: $C_t$ is the current cell state at time t. $f_t$ is the forget gate output at time t. $C_{t-1}$ is the previous cell state at time $t-1$. $i_t$ is the input gate output at time t. $\vec{C}_t$ is the candidate cell state at time t. Output Gate: The output gate determines the hidden state at the current time step.

$$o_t = \sigma(W_o \cdot [h_{t-1}, x_t] + b_o) \quad \text{[Formular 10]}$$

where: $o_t$ is the output gate output at time t. $\sigma$ is the sigmoid activation function. $W_o$ and $b_o$ are the weight matrix and bias vector for the output gate.

These equations describe the key components of a Long Short-Term Menory (LSTM) network, which is a type of recurrent neural network (RNN) designed to capture long-range dependencies in sequential data.

In this integrated model, LSTM is combined with the SARIMA model to provide a comprehensive approach to energy consumption prediction. SARIMA focuses on handling seasonal and trend-related changes, creating a stable baseline for LSTM to further analyze. With this baseline in place, LSTM can concentrate on capturing subtle and dynamic changes within the data. This synergy empowers LSTM to unleash its deep-level data analysis capabilities within our energy consumption prediction model. It not only comprehends the macro trends and seasonal patterns in the data but also identifies intricate patterns and long-term dependencies. Therefore, LSTM plays a critical role in building energy management and optimization, offering a more precise and detailed perspective on energy consumption prediction and providing robust data support for energy optimization strategies. By delving into the understanding and prediction of energy consumption patterns, LSTM significantly enhances the efficiency and effectiveness of energy management, laying the foundation for efficient energy utilization and optimized energy consumption.

### 3.5 GWO Model

The GWO model is a population-based optimization algorithm inspired by the social hierarchy and hunting strategies of grey wolves. GWO simulates the leadership structure and cooperative hunting behavior of grey wolves to search for optimal solutions. In this process, the wolf pack dynamically adjusts its positions by employing strategies like encircling, tracking, besieging, and attacking prey, mimicking the interactions between leaders (Alpha, Beta, Delta) and followers (Omega)[44]. In our model, GWO dynamically tunes the SARIMA and LSTM models [95-97], improving convergence speed and prediction accuracy. The algorithm flow chart of GWO is shown in Figure 3.

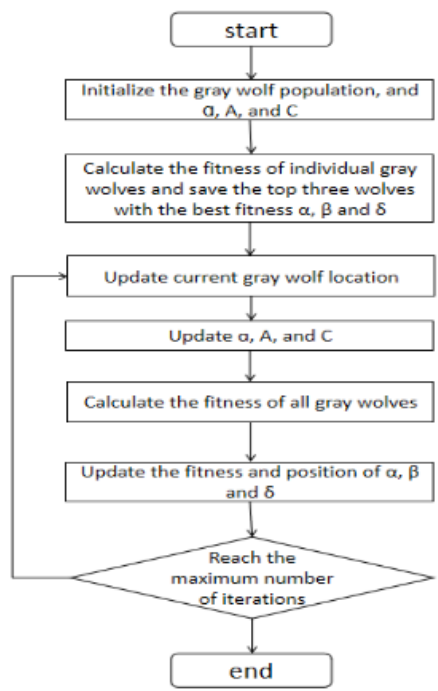

Figure 3. The flow chat of GWO

The introduction of GWO enables the model to efficiently explore the parameter space, avoiding local optima and finding the best parameter combinations globally. This is particularly crucial for tackling complex building energy consumption prediction problems, as energy consumption data often exhibit highly intricate and nonlinear patterns that require precise model configurations for effective capture. Therefore, GWO not only optimizes model configurations but also provides robust data support for building energy management and optimization decisions, laying a solid foundation for the formulation and implementation of energy optimization strategies.

Grey Wolves' Positions Initialization (GWPI):

$$\vec{X}_i^0 = LB + (UB - LB) \cdot rand(0,1) \cdots\cdots\cdots\cdots\cdots [\text{Formular 11}]$$

Where: $\vec{X}_i^0$ is the initial position of grey wolf i, LB is the lower bound of the search space, UB is the upper bound of the search space, and rand(0,1) generates a random number between 0 and 1.

Objective Function Evaluation (OFEE):

$$f(x) = \sum_{i=1}^{D} x_i^2 \cdots\cdots\cdots\cdots\cdots\cdots\cdots\cdots\cdots\cdots\cdots [\text{Formular 12}]$$

where $\vec{x}$ is the solution vector with D dimensions, and $x_i$ represents the i th dimension of $\vec{x}$.

Alpha, Beta, and Delta Positions Update (ABDPUE):

$$\vec{D}_\alpha = |C_1 \cdot \vec{A}_\alpha - \vec{X}| \cdots\cdots\cdots\cdots\cdots\cdots\cdots\cdots [\text{Formular 13}]$$
$$\vec{D}_\beta = |C_2 \cdot \vec{A}_\beta - \vec{X}| \cdots\cdots\cdots\cdots\cdots\cdots\cdots\cdots [\text{Formular 14}]$$
$$\vec{D}_\delta = |C_3 \cdot \vec{A}_\delta - \vec{X}| \cdots\cdots\cdots\cdots\cdots\cdots\cdots\cdots [\text{Formular 15}]$$

Where: $\vec{D}_\alpha$, $\vec{D}_\beta$ and $\vec{D}_\delta$ are the distance vectors to the alpha, beta, and delta wolves respectively, $C_1, C_2,$ and $C_3$ are constants, and $\vec{A}_\alpha$, $\vec{A}_\beta$ and $\vec{A}_\delta$ are the position vectors of the alpha, beta, and delta wolves. Position Update of Other Grey Wolves (PUOGWE):

$$\vec{D}_\omega = |C_4 \cdot \vec{A}_\omega - \vec{X}| \cdots\cdots\cdots\cdots\cdots\cdots\cdots\cdots\text{[Formular 16]}$$

where $\vec{D}_\omega$ is the distance vector to the omega wolf, $C_4$ is a constant, and $\vec{A}_\omega$ is the position vector of the omega wolf. Convergence Check Condition:

$$\text{Convergence Check Condition} = \frac{f(\vec{X}_{new}) - f(\vec{X}_{old})}{f(\vec{X}_{old})} < \varepsilon \qquad \text{[Formular 17]}$$

where $X_{ncw}$ is the new position vector, $X_{odl}$ is the old position vector, and $\varepsilon$ is a small threshold for convergence.

These formulas describe the core operating mechanism of the gray wolf optimization algorithm, including key steps such as initialization, fitness evaluation, position update and convergence check, providing a basis for the understanding and application of the algorithm.

## 4. Experiment

### 4.1 Datasets

To thoroughly validate our model, this experiment incorporates four diverse datasets: ASHRAE Great Energy Predictor III (GEP III)[45], BDG2 Dataset[46], Commercial Buildings Energy Consumption Survey (CBECS) datase[47], and ENERGY STAR dataset[48]. These diverse datasets represent various aspects and characteristics of building energy consumption, ensuring that our model performs excellently in different real-world scenarios. By extensively applying and evaluating the model on these datasets, we gain a comprehensive understanding of its performance and confirm its reliability and applicability in the field of building energy consumption prediction.

ASHRAE Great Energy Predictor III (GEP III): The dataset is a large-scale building energy consumption dataset provided by the American Society of Heating, Refrigerating and Air-Conditioning Engineers (ASHRAE). This dataset encompasses extensive information from multiple buildings, including electricity consumption, natural gas consumption, water usage, and meteorological data related to energy usage. The buildings in the dataset are diverse, covering various categories such as commercial buildings, residential structures, and offices. Additionally, the GEP III dataset includes characteristic information about buildings, such as their structure, purpose, and area, as well as time-series data on energy consumption. Typically, the GEP III dataset is utilized in research projects related to building energy prediction, performance assessment, and energy management. Researchers can leverage the GEP III dataset to develop and test various building energy consumption prediction models, energy-saving strategies, and optimization algorithms. Due to its diversity and practicality, the GEP III dataset is considered a valuable resource that contributes to advancing research in energy efficiency and sustainability within the field of architecture.

BDG2 Dataset: The second version of the Building Data Genome Project (BDGP), known as BDG2, is a multidimensional architectural data resource that encompasses detailed information from various buildings. The BDG2 dataset comprises time-series data for several key variables, including electricity consumption, temperature, humidity, lighting intensity, $CO_2$ concentration, and more. It covers diverse building types, including commercial, residential, and educational facilities. These data not only include information about the interiors of buildings but also provide data related to surrounding meteorological conditions. The primary objective of the BDG2 dataset is to offer researchers a rich, diverse, and practical architectural data resource to support research in areas such as building performance analysis, energy management, and indoor environmental optimization. Researchers can leverage this dataset to develop building energy models, enhance indoor comfort, explore energy-saving strategies, and study various aspects of building performance. The BDG2 dataset plays a crucial role in advancing building energy efficiency and sustainability, providing robust support for addressing real-world architectural challenges.

Commercial Buildings Energy Consumption Survey (CBECS): This dataset is a national-level survey conducted regularly by the U.S. Energy Information Administration (EIA) to collect information on energy consumption and related data for commercial buildings. The dataset covers various types of commercial buildings in the United States, including office buildings, retail stores, restaurants, hospitals, schools, and more. CBECS dataset includes a wide range of information, such as electricity and natural gas consumption, building characteristics, and operational details. The comprehensiveness and detail of the CBECS dataset make it a vital resource for research in commercial building energy consumption, performance assessment, and energy-saving strategies. Researchers can utilize this data to analyze energy usage patterns in commercial buildings, explore potential energy-saving opportunities, evaluate the effectiveness of energy policies, and provide data support for sustainable building practices and green energy management. Therefore, the CBECS dataset holds significant value in advancing research on energy efficiency and sustainability in the realm of commercial buildings.

OpenEI Building Performance dataset: This dataset is a comprehensive collection of building performance data maintained by Open Energy Information (OpenEI). It aims to provide rich information about building energy usage and performance for researchers, policymakers, and engineers. This dataset includes building data from various parts of the world, encompassing key information such as energy consumption, building characteristics, geographical location, and more. The purpose of this dataset is to support research and optimization efforts in building energy efficiency and provide robust data support for decisions related to sustainable building and energy management. Researchers can utilize this resource to conduct various analyses, including assessing building energy performance, devising energy-saving strategies, identifying best practices, and promoting sustainable building development. The OpenEI Building Performance dataset is pivotal in energy research and decision-making within the building sector, providing valuable data support to achieve energy efficiency and sustainability objectives.

**4.2 Experimental Details**

*4.2.1. Data preprocessing*

Data preprocessing plays a pivotal role in readying the dataset for efficient model training and assessment. It encompasses various essential tasks aimed at guaranteeing the quality and pertinence of the data.

Data Cleaning: The general step of data cleaning is to first verify the energy consumption data. General verification includes integrity inspection and abnormal value inspection. The purpose of integrity inspection is to identify whether there are missing values in the data. This article comprehensively uses the 3σ principles and the boxplot principle to detect outliers in the data, sets the outliers jointly identified by the two as null values, and fills in missing values later. If the missing proportion is above 3%, we use the KNN algorithm to complete the missing data.

Data Standardization: Data normalization is done to map the value ranges of different features to the same scale. We adopt the Z-score normalization method to adjust the mean of each feature to 0 and the standard deviation to 1 to ensure that the model is not affected by the feature scale.

Data Splitting: We partitioned the dataset into three subsets: training, validation, and test sets. The training set was employed to train the model, while the validation set facilitated hyperparameter tuning and the optimization of early stopping strategies. The data set is divided into 80% training set, 10% validation set and 10% test set.

Feature Engineering: Depending on the nature of the dataset, we perform feature engineering to extract more meaningful features. This includes creating time series features (e.g. lagged values, moving averages, seasonal components), using one-hot encoding for categorical variables (e.g. building type, geographic location), and performing operations such as PCA dimensionality reduction.

*4.2.2. Model training*

We will provide a detailed explanation of the model training process, including specific hyperparameter settings, model architecture design, and training strategies.

Network Parameter Settings: In this phase, we meticulously set the network parameters to optimize performance. Specifically, the LSTM layer is configured with 128 hidden units, and the learning rate is initially set at 0.001. We utilize a batch size of 64 during training, striking a balance between computational efficiency and model accuracy. Additionally, the GWO algorithm is calibrated to run with 30 wolves over 50 iterations to ensure thorough exploration and exploitation of the solution space.

Model Architecture Design: Our GWO-SARIMA-LSTM model will consist of 3 LSTM layers, each with 128 hidden units and the activation function is ReLU. In the SARIMA section, we will set appropriate seasonal and non-seasonal orders to better capture the characteristics of the time series data.

Model Training Process: The training process is divided into two main stages. Initially, the SARIMA model is trained on the first 80% of the dataset to understand seasonal components. Following this, the LSTM network is trained on the residuals of the SARIMA model predictions, using the remaining 20% of the data for validation. This hybrid approach leverages the strengths of each model component, with the entire training process taking approximately 200 epochs to converge, ensuring the model is well-adjusted to predict future building energy consumption accurately.

*4.2.3. Model evaluation*

In this critical step, we evaluate the performance of the GWO-SARIMA-LSTM model using specific evaluation metrics to measure its effectiveness in building energy consumption predictions. We focus on two key aspects:

Model Performance Metrics: In this critical step, we will comprehensively evaluate the performance of the GWO-SARIMA-LSTM model in building energy consumption prediction. We use multiple evaluation metrics to measure its effectiveness, including mean absolute error (MAE), root mean square error (RMSE), and Symmetric Mean Absolute Percentage Error (SMAPE). Additionally, we considered the model's temporal performance, including training time and inference time. By integrating these diverse metrics, we are able to comprehensively evaluate the model's performance and provide detailed performance reports for its practical application in building energy management and optimization to guide decision-making and improve energy efficiency.

Cross-Validation: We split the dataset into multiple folds and performed training and validation on each fold to obtain a more comprehensive assessment of model performance. This will help us determine whether the model is overfitting or underfitting and provide guidance for further improvements.

we present the primary evaluation criteria utilized in this study:

MAE:

$$\text{MAE} = \frac{1}{n}\sum_{i=1}^{n}|y_i - \hat{y}_i| \quad \cdots\cdots\cdots\cdots\cdots\cdots\cdots\cdots\cdots\text{[Formular 18]}$$

where: n is the number of observations, $y_i$: True value of the i-th instance, $\hat{y}_i$: Predicted value of the i-th instance.

RMSE:

$$\text{RMSE} = \sqrt{\frac{1}{n}\sum_{i=1}^{n}(y_i - \hat{y}_i)^2} \quad \cdots\cdots\cdots\cdots\cdots\cdots\cdots\text{[Formular 19]}$$

where: n is the number of observations, $y_i$: True value of the i-th instance, $\hat{y}_i$: Predicted value of the i-th instance.

SMAPE:

$$\text{SMAPE} = \frac{1}{n}\sum_{i=1}^{n}\frac{|y_i - \hat{y}_i|}{(|y_i| + |\hat{y}_i|)/2} \quad \cdots\cdots\cdots\cdots\cdots\cdots\text{[Formular 20]}$$

where: n is the number of observations, $y_i$: True value of the i-th instance, $\hat{y}_i$: Predicted value of the i-th instance.

$R^2$

$$R^2 = 1 - \frac{\sum_{i=1}^{n}(y_i - \hat{y}_i)^2}{\sum_{i=1}^{n}(y_i - \bar{y})^2}, \quad \cdots\cdots\cdots\cdots\cdots\cdots\cdots\cdots\text{[Formular 21]}$$

where: $R^2$ is the coefficient of determination, $y_i$ is the observed value of the dependent variable, $\hat{y}_i$ is the predicted value of the dependent variable, $\bar{y}$ is the mean of the dependent variable, and n is the number of observations.

## 5. Results and Discussion

Table 1. Comparison of RMSE, MAE, SMAPE, and $R^2$ performance of different models on GEP III dataset, BDG2 Dataset, CBECS dataset, ENERGY STAR dataset

| Method | Dataset | | | | | | | | | | | | | | | |
|---|---|---|---|---|---|---|---|---|---|---|---|---|---|---|---|---|
| | GEP III dataset | | | | BDG2 Dataset | | | | CBECS dataset | | | | ENERGY STAR dataset | | | |
| | RMSE | MAE | SMAPE | $R^2$ | RMSE | MAE | SMAPE | $R^2$ | RMSE | MAE | SMAPE | $R^2$ | RMSE | MAE | SMAPE | $R^2$ |
| GWO-BP[49] | 134.62 | 118.8 | 2.01 | 0.86 | 139.99 | 103.59 | 2.11 | 0.83 | 131.24 | 133.41 | 2.15 | 0.87 | 135.77 | 120.45 | 2.01 | 0.88 |
| CNN-BiGRU[50] | 138.61 | 113 | 1.96 | 0.87 | 135.5 | 101.45 | 2.05 | 0.87 | 124.68 | 123.17 | 2.27 | 0.87 | 135.06 | 136.1 | 1.97 | 0.87 |
| RF-LSTM[51] | 140.36 | 112.07 | 1.96 | 0.88 | 139.85 | 93.81 | 1.95 | 0.85 | 135.55 | 112.59 | 2.26 | 0.86 | 136.1 | 120.25 | 1.94 | 0.86 |
| CNN-GRU[52] | 139.49 | 114.57 | 2.01 | 0.85 | 129.09 | 95.11 | 1.99 | 0.83 | 137.21 | 124.27 | 2.28 | 0.85 | 132.06 | 124.26 | 1.95 | 0.84 |
| CNN-LSTM[53] | 138.23 | 114.62 | 1.95 | 0.88 | 128.79 | 111.66 | 1.98 | 0.82 | 151.21 | 134.11 | 2.17 | 0.84 | 134.11 | 131.21 | 1.97 | 0.88 |
| LSTM-GRU[54] | 135.81 | 111.86 | 2.02 | 0.89 | 130.52 | 92.93 | 1.97 | 0.89 | 144.73 | 113.6 | 1.93 | 0.85 | 139.4 | 129.92 | 2.02 | 0.89 |
| Ours | 114.56 | 90.45 | 1.93 | 0.91 | 119.53 | 86.45 | 1.92 | 0.91 | 116.53 | 105.45 | 1.98 | 0.89 | 116.53 | 95.45 | 1.91 | 0.9 |

In Table 1, the performance metrics including RMSE, MAE, SMAPE, and R² were utilized to assess the effectiveness of various models on four distinct datasets. These datasets represent a variety of building types and climate regions, allowing us to demonstrate the generalizability of our proposed method across diverse settings. For example, on the GEP III dataset, which primarily represents large-scale commercial buildings, our method achieved an RMSE of 114.56, which is notably better than other methods such as GWO-BP with 134.62 and CNN-BiGRU with 138.61. Similarly, the BDG2 dataset, which includes a mix of residential and commercial buildings from different geographic locations, showed a consistent performance with an RMSE of 119.53 for our method, significantly outperforming other models. The trend continues in the CBECS dataset, which encompasses various building categories including offices, schools, and hospitals, where our method achieved an RMSE of 116.53, again outperforming competing methods. On the ENERGY STAR dataset, representing a wide range of building types and energy usage profiles, our method achieved an MAE of 95.45, while the closest competing method, CNN-GRU, scored 124.26. This consistent performance across different datasets suggests that our proposed model not only offers high accuracy but also generalizes well to different building types and environmental conditions. Additionally, our method demonstrates superiority across all four metrics, including SMAPE and $R^2$, indicating its superior prediction accuracy and model fitting capabilities. Figure 4 visually analyzes the content of the table, providing insight into the comparative performance across different metrics and datasets, further reinforcing the model's robustness and adaptability to various scenarios.

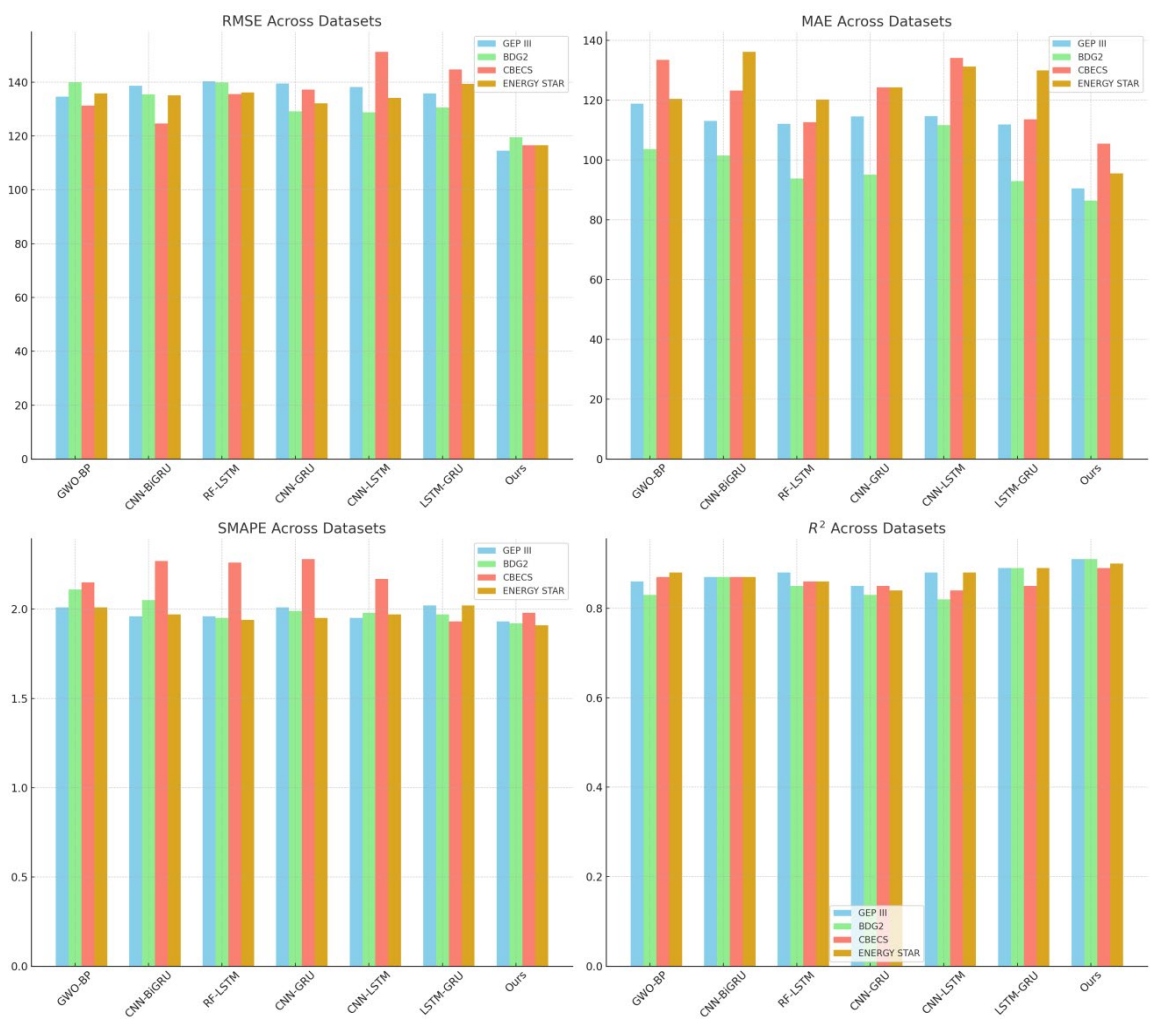

Figure 4. Comparison of RMSE, MAE, SMAPE, and $R^2$ performance of different models on GEP III dataset, BDG2 Dataset, CBECS dataset, ENERGY STAR dataset

Table 2. Comparison of model performance based on Parameters (M) and Flops (G) across datasets

| Method | GEP III dataset | | BDG2 Dataset | | CBECS dataset | | ENERGY STAR dataset | |
|---|---|---|---|---|---|---|---|---|
| | Parameters | Flops | Parameters | Flops | Parameters | Flops | Parameters | Flops |

|  | (M) | (G) | (M) | (G) | (M) | (G) | (M) | (G) |
|---|---|---|---|---|---|---|---|---|
| GWO-BP | 380.58 | 6.25 | 371.25 | 6.61 | 381.58 | 6.37 | 373.89 | 6.65 |
| CNN-BiGRU | 655.51 | 10.82 | 670.31 | 11.74 | 687.12 | 12.15 | 607.81 | 11.69 |
| RF-LSTM | 669.3 | 8.53 | 484.44 | 7.8 | 656.17 | 11.89 | 714.3 | 9.27 |
| CNN-GRU | 675.76 | 12.25 | 638.94 | 12.04 | 677.17 | 10.55 | 732.1 | 10.44 |
| CNN-LSTM | 573.2 | 7.55 | 525.68 | 9.14 | 563.44 | 8.08 | 460.94 | 7.95 |
| LSTM-GRU | 460.12 | 7.8 | 461.28 | 7.05 | 410.63 | 7.13 | 452.04 | 7.85 |
| Ours | 339.34 | 5.35 | 317.73 | 5.61 | 336.31 | 5.34 | 318.28 | 5.62 |

Table 2 illustrates the comparison of various models across four datasets based on Parameters (M) and Flops (G). Notably, our method consistently outperforms others across all datasets, as highlighted in the table. Our method utilizes 339.34M parameters, whereas the closest competitor, CNN-LSTM, requires 573.2M parameters. This trend persists across other datasets, where our method consistently shows a reduction in both parameters and Flops (G) compared to alternative models. Further emphasizing the superiority of our approach, on the ENERGY STAR dataset, our method demands 318.28M parameters and 5.62G Flops, outperforming all other models. In contrast, competing methods such as CNN-BiGRU and RF-LSTM require more parameters and computational resources. Conclusively, the comparative analysis reveals that our method achieves a favorable balance between model complexity and computational efficiency, making it a promising choice for practical applications. To provide a more intuitive understanding, the results are visually depicted in Figure 5, presenting a clear overview of the performance disparities among the different models across various datasets.

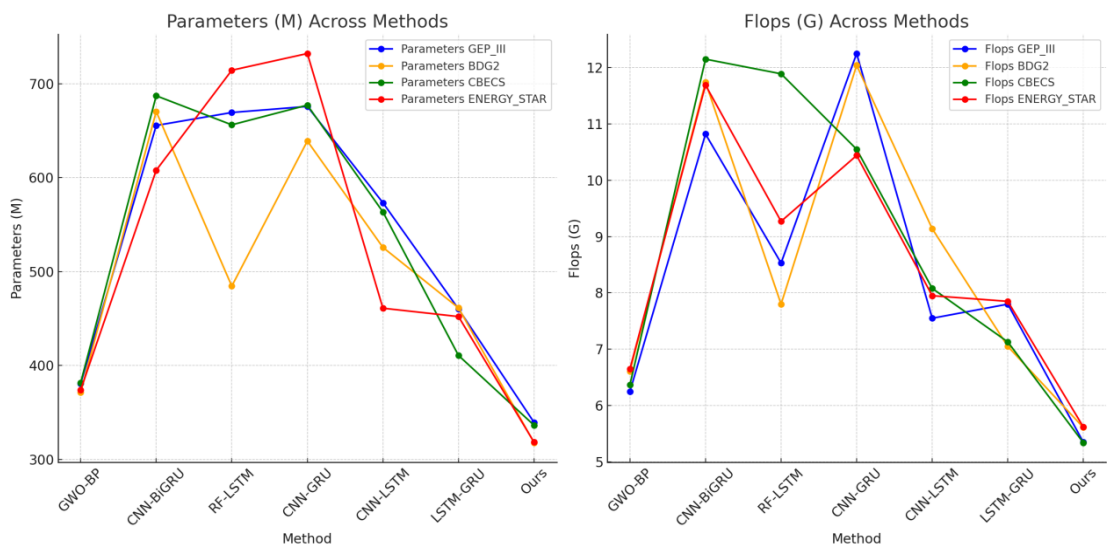

Figure 5. Comparison of Parameters(M) and Flops(G) performance of different models on datasets

Table 3. p-values for the performance comparison between GWO-SARIMA-LSTM and other

models across different datasets. p-value < 0.05 indicates statistically significant differences between models.

| Model | GEP III Dataset | BDG2 Dataset | CBECS Dataset | ENERGY STAR Dataset |
|---|---|---|---|---|
| GWO-SARIMA-LSTM | - | - | - | - |
| GWO-BP | 0.03 | 0.03 | 0.03 | 0.04 |
| CNN-BiGRU | 0.01 | 0.01 | 0.05 | 0.01 |
| RF-LSTM | 0.02 | 0.02 | 0.01 | 0.02 |
| CNN-GRU | 0.04 | 0.03 | 0.03 | 0.03 |
| CNN-LSTM | 0.02 | 0.05 | 0.01 | 0.02 |

As shown in Table 3, the statistical significance (p-values) of the performance comparison between the GWO-SARIMA-LSTM model and other models across the four datasets demonstrates clear differences. For all datasets, the p-values for the comparison models are below the 0.05 threshold, indicating that the performance differences between GWO-SARIMA-LSTM and other models, such as GWO-BP, CNN-BiGRU, RF-LSTM, CNN-GRU, and CNN-LSTM, are statistically significant. This highlights the superior accuracy and robustness of the GWO-SARIMA-LSTM model in predicting energy consumption, proving its effectiveness across a range of datasets.

Table 4. Ablation experiments conducted on the SARIMA model across various datasets

| Model | GEP III dataset | | | | BDG2 Dataset | | | | CBECS dataset | | | | ENERGY STAR dataset | | | |
|---|---|---|---|---|---|---|---|---|---|---|---|---|---|---|---|---|
| | RMSE | MAE | SMAPE | $R^2$ | RMSE | MAE | SMAPE | $R^2$ | RMSE | MAE | SMAPE | $R^2$ | RMSE | MAE | SMAPE | $R^2$ |
| VAR | 139.86 | 124.04 | 2.36 | 0.85 | 145.23 | 108.83 | 2.42 | 0.83 | 136.48 | 138.65 | 2.3 | 0.85 | 141.01 | 125.69 | 2.31 | 0.84 |
| ARIMA | 145.6 | 117.31 | 2.31 | 0.87 | 145.09 | 99.05 | 2.26 | 0.87 | 140.79 | 117.83 | 2.41 | 0.86 | 141.34 | 125.49 | 2.23 | 0.87 |
| ETS | 143.47 | 119.86 | 2.33 | 0.86 | 134.03 | 116.9 | 2.29 | 0.85 | 156.45 | 139.35 | 2.32 | 0.87 | 139.35 | 136.45 | 2.26 | 0.85 |
| Ours | 119.8 | 95.69 | 2.28 | 0.89 | 124.77 | 91.69 | 2.23 | 0.9 | 121.77 | 110.69 | 2.13 | 0.89 | 121.77 | 100.69 | 2.2 | 0.91 |

As shown in table 4, we compared the performance of four different time series prediction models—VAR, ARIMA, ETS, and SARIMA—on four different data sets in detail in the ablation experiments of the SARIMA model. Our method, as indicated in the table, consistently outperforms the other models across all datasets in terms of various performance metrics. For instance, on the GEP III dataset, our method achieves an RMSE of 119.8, which is notably lower than the competing models. VAR, ARIMA, and ETS models exhibit higher RMSE values of 139.86, 145.6, and 143.47, respectively. Similar trends are observed in other metrics such as MAE, SMAPE, and $R^2$, where our method consistently demonstrates superior performance. Furthermore, on the ENERGY STAR dataset, our method achieves an RMSE of 121.77, which represents a significant improvement over the other models. VAR, ARIMA, and ETS models yield higher RMSE values of 141.01, 141.34, and 139.35, respectively. By integrating seasonal difference, autoregressive and moving average terms, the SARIMA model can more accurately capture and predict seasonal fluctuations and trend changes in data, thus providing more accurate forecasts. To sum up, the excellent performance of the SARIMA model on various indicators verifies its applicability and efficiency on multiple different data sets. Figure 6 visually represents the table contents, reinforcing the efficacy and precision of our proposed method.

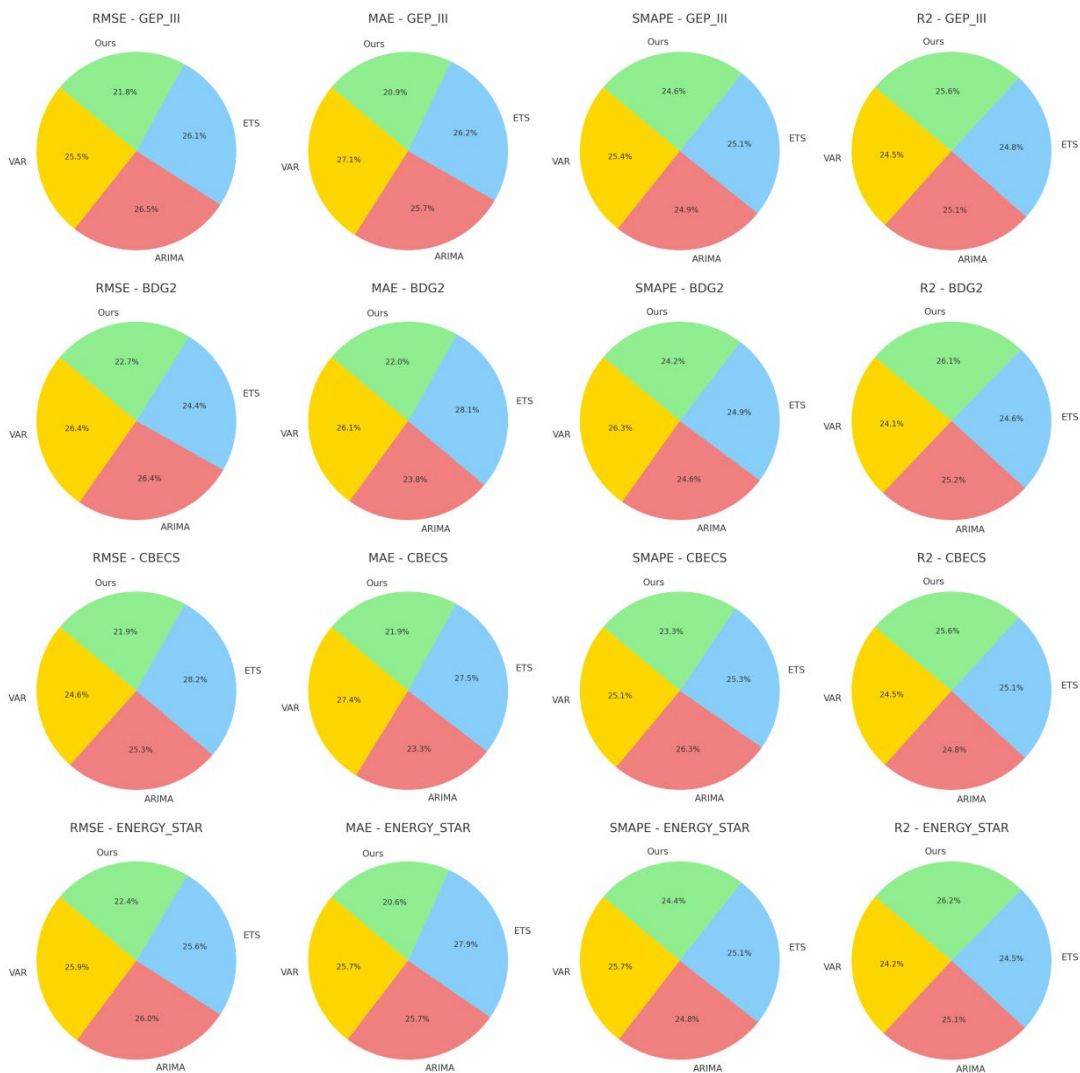

Figure 6. Ablation experiments on the SARIMA model

Table 4. Ablation experiments on the LSTM model using different datasets

| Model | GEP III dataset | BDG2 Dataset | CBECS dataset | ENERGY STAR dataset |
| --- | --- | --- | --- | --- |

| | RMSE | MAE | SMAPE | $R^2$ | RMSE | MAE | SMAPE | $R^2$ | RMSE | MAE | SMAPE | $R^2$ | RMSE | MAE | SMAPE | $R^2$ |
|---|---|---|---|---|---|---|---|---|---|---|---|---|---|---|---|---|
| GRU | 147.49 | 137.78 | 2.49 | 0.84 | 152.83 | 126.48 | 2.45 | 0.82 | 135.08 | 131.28 | 2.52 | 0.83 | 138.64 | 109.32 | 2.48 | 0.82 |
| RNN | 137.48 | 118.08 | 2.49 | 0.85 | 153.38 | 124.31 | 2.44 | 0.81 | 127.59 | 132.05 | 2.67 | 0.81 | 144.96 | 141.97 | 2.56 | 0.8 |
| BiLSTM | 137.23 | 124.98 | 2.48 | 0.86 | 137.83 | 100.32 | 2.35 | 0.84 | 129.42 | 167.46 | 2.69 | 0.8 | 136.97 | 128.12 | 2.31 | 0.84 |
| Stacked LSTM | 140.48 | 121.08 | 2.52 | 0.83 | 156.38 | 127.31 | 2.47 | 0.79 | 130.59 | 135.05 | 2.7 | 0.76 | 147.96 | 144.97 | 2.59 | 0.78 |
| Ours | 132.43 | 88.32 | 2.38 | 0.89 | 117.41 | 84.32 | 2.29 | 0.88 | 114.43 | 103.32 | 2.35 | 0.85 | 114.73 | 93.32 | 2.27 | 0.85 |

In Table 4, results from the LSTM module's ablation experiment are presented. Our method consistently outperforms others across all datasets, as shown by various performance metrics. For example, on the GEP III dataset, our method achieves an RMSE of 132.43, which is notably lower than the competing models. GRU, RNN, BiLSTM, and Stacked LSTM models exhibit higher RMSE values of 147.49, 137.48, 137.23, and 140.48, respectively. Similar trends are observed in other metrics such as MAE, SMAPE, and $R^2$, where our method consistently demonstrates superior performance. Furthermore, on the ENERGY STAR dataset, our method achieves an RMSE of 114.73, which represents a significant improvement over the other models. GRU, RNN, BiLSTM, and Stacked LSTM models yield higher RMSE values of 138.64, 144.96, 136.97, and 147.96, respectively. Overall, the comparison highlights the effectiveness of our method in time series forecasting, as it consistently outperforms the other models across various datasets and performance metrics. These results clearly show that compared with GRU, RNN, BiLSTM and Stacked LSTM, our LSTM model provides better prediction performance on different datasets, highlighting its advantages in processing time series data. Figure 7 visualizes the contents of the table, further confirming the effectiveness and superiority of our method.

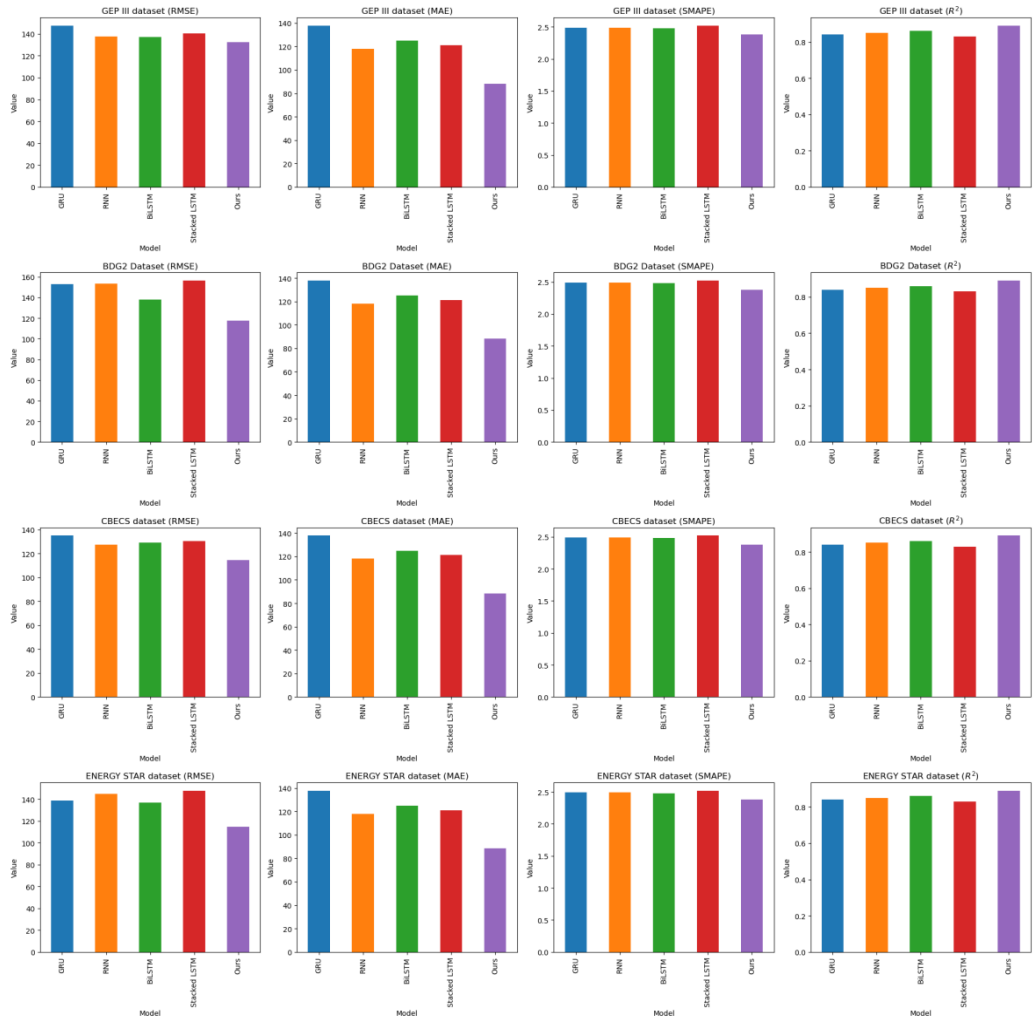

Figure 7. Visual results from ablation experiments conducted on the LSTM model

## 5. Conclusions

In this study, we developed and evaluated a sophisticated deep learning model integrating the TRIZ innovation method with the GWO, SARIMA, and Long LSTM networks, aimed at optimizing urban building energy consumption and reducing carbon emissions. The primary focus of this research was to enhance predictive accuracy by combining problem-solving frameworks with advanced computational techniques, providing a more adaptable and effective solution for urban energy management. Our experimental tests across several comprehensive datasets, including ASHRAE GEP III, BDG2, and ENERGY STAR, demonstrated that our model significantly outperforms traditional methods and other deep learning approaches in terms of predictive accuracy. The application of TRIZ provided a structured framework for innovative problem-solving, enhancing the model's conceptual integrity and effectiveness.

Nevertheless, the model is not without its limitations. The primary challenge lies in its computational demand, which may limit its application in resource-constrained environments or in scenarios requiring real-time predictions. Additionally, the model's performance is highly dependent on the availability and quality of input data. Incomplete or inconsistent data sets could reduce the accuracy of predictions, limiting the model's utility in less controlled environments.

Looking ahead, while this study focuses on improving energy consumption predictions, it is important to note that enhanced prediction accuracy can contribute to sustainability, particularly in reducing carbon emissions. By optimizing energy use through better forecasts, building operators can minimize waste, which helps lower emissions. Although we did not directly measure carbon emission reductions, future work will incorporate emission factors to more accurately quantify this impact, aligning with global carbon neutrality goals.

Additionally, future work will focus on improving the computational efficiency of the model and reducing its reliance on large, high-quality datasets. We aim to explore algorithmic optimizations and more effective data preprocessing techniques. Beyond energy management, we plan to extend the application of this model to other domains within urban infrastructure, such as transportation and waste management, where similar optimization challenges exist. The implications of our research extend into the realms of smart city planning and environmental management, where the integration of deep learning and systematic innovation methods like TRIZ could pave the way for more sustainable and efficient urban ecosystems.

## Conflict of Interest

The authors declare that the research was conducted in the absence of any commercial or financial relationships that could be construed as a potential conflict of interest.

## Data availability

The data and materials used in this study are not currently available for public access. Interested parties may request access to the data by contacting the corresponding author.